\documentclass[runningheads]{llncs}

\usepackage[T1]{fontenc}
\usepackage{graphicx}
\usepackage{amsfonts, amssymb}
\begin{document}

\title{Bridging Scientific Heritage: An Arabic--Russian Parallel Corpus and LLM Benchmark for Sustainable Knowledge Transfer}

\titlerunning{LLM Benchmark for Arabic--Russian Translation}

\author{M. K. Arabov\inst{1}\orcidID{0000-0003-2525-1183}}

\authorrunning{M. K. Arabov}

\institute{Kazan Federal University,
Institute of Computational Mathematics and Information Technologies,
Department of Data Analysis,
Kazan, Russia\\
\email{MKArabov@kpfu.ru}}

\maketitle     
\begin{abstract}
Russian and Arabic are among the major languages of scientific communication. Language barriers impede the exchange of research results between these communities, which affects international collaboration and the progress of sustainability-related research.

We present a benchmark for Arabic--Russian scientific translation. The benchmark includes a hybrid parallel corpus of about 27,000 sentence pairs, compiled from scientific abstracts and general-domain texts (religion, news, conversations). We fine-tune three multilingual language models --- mT5-base (580M parameters), NLLB-200-distilled-1.3B (1.3B), and Qwen2.5-7B-Instruct (7B) --- using LoRA with ranks 8, 16, 32, and 64.

The Qwen2.5-7B model with QLoRA (rank 8) yields BLEU 23.15, chrF 43.89, BERTScore 0.906, and COMET 0.758. These are +4.36 BLEU and +0.051 COMET above the zero-shot baseline. Few-shot prompting with three examples does not improve performance, indicating that domain-specific fine-tuning is required.

We release the models, the corpus, and the evaluation code. By lowering the language barrier for scientific texts, the work enables knowledge exchange between Arabic-speaking and Russian-speaking researchers. It contributes to sustainable partnerships (UN SDG 17) and innovation infrastructure (SDG 9), aligning with the conference's focus on technology-driven sustainable development.

\end{abstract}

\section{Introduction}

Russian and Arabic are among the major languages of scientific communication. Russian-language scholarship has contributed foundational advances across physics, mathematics, engineering, and medicine --- from the work of Lobachevsky and Mendeleev to modern computational mathematics and space research. The Arabic-speaking world has produced a rich intellectual tradition spanning astronomy, medicine, and algebra, with contemporary research institutions across the Gulf region contributing to sustainability science, renewable energy, and water management \cite{HadjAmeur:2020:Survey}.

Despite these complementary scientific traditions, a language barrier continues to impede knowledge exchange between these communities. While Arabic--English translation has received substantial research attention and resource development \cite{Al-Matham:2025:BALSAM}, \cite{Alzubaidi:2025:Survey}, the Arabic--Russian pair remains under-resourced. This gap is relevant given the academic ties between these regions and shared challenges in sustainable development, energy transition, and climate adaptation. Facilitating knowledge transfer between these communities can help researchers leverage complementary expertise, avoid duplication, and accelerate progress in areas such as water management, renewable energy, and climate resilience.

Recent advances in multilingual large language models \cite{Hu:2022:LoRA}, \cite{Dettmers:2024:QLoRA} have improved machine translation performance, yet their effectiveness for Arabic--Russian scientific translation has not been systematically evaluated. Existing parallel corpora for this language pair are predominantly focused on religious texts, conversational phrases, or general-domain content, and lack the specialized terminology required for scientific and technical translation. The proliferation of open-source architectures --- encoder-decoder models (NLLB, mT5) and decoder-only instruction-tuned models (Qwen) --- raises the question of which approaches are most suitable for low-resource scientific translation \cite{Song:2026:SmallLM}.

To address these gaps, we present a benchmark for Arabic--Russian scientific translation with the following contributions:

\begin{enumerate}
\item A hybrid parallel corpus of approximately 27,000 training examples, constructed from scientific abstracts and six general-domain sources (religion, news, conversations, dictionaries, Bible, and Tatoeba), ensuring both domain adaptability and linguistic diversity.
\item A systematic evaluation of three multilingual large language models of varying sizes --- mT5-base (580M), NLLB-200-distilled-1.3B (1.3B), and Qwen2.5-7B-Instruct (7B) --- fine-tuned with Low-Rank Adaptation (LoRA) \cite{Hu:2022:LoRA} and QLoRA \cite{Dettmers:2024:QLoRA} across ranks 8, 16, 32, and 64.
\item A rigorous comparative analysis using automatic metrics including BLEU \cite{Papineni:2002:BLEU}, chrF \cite{Popovic:2015:chrF}, BERTScore \cite{Zhang:2020:BERTScore}, and COMET \cite{Rei:2020:COMET}, establishing a reproducible baseline for future research.
\item Public release of all fine-tuned models, the evaluation pipeline, and the curated corpus to facilitate reproducible research and practical deployment in the Arabic--Russian translation community.
\end{enumerate}

The results show that the Qwen2.5-7B-Instruct model fine-tuned with QLoRA (rank 8) achieves a BLEU score of 23.15 and COMET of 0.758, outperforming zero-shot baselines by +4.36 BLEU and +0.051 COMET. Few-shot prompting with three in-context examples does not improve performance, indicating that domain-specific fine-tuning is required. Smaller encoder-decoder models such as mT5-base, even with parameter-efficient fine-tuning, do not achieve acceptable translation quality for this language pair, suggesting a minimum model capacity requirement for scientific Arabic--Russian translation.

By enabling high-quality, automated translation of scientific content, our work addresses a knowledge transfer barrier between Arabic-speaking and Russian-speaking research communities. This transfer is particularly relevant for sustainability science, where challenges such as desertification, water scarcity, and renewable energy integration require international cooperation and the exchange of research findings across linguistic boundaries. Through systematic benchmarking of LLMs for Arabic--Russian scientific translation, this study provides a replicable framework for reducing linguistic barriers and promoting sustainable knowledge exchange across regions with complementary scientific expertise.

\section{Related Work}

This section reviews the relevant literature across three interconnected areas: Arabic machine translation and parallel corpora, multilingual large language models and parameter-efficient fine-tuning, and cross-script NLP with translation evaluation. We examine the contributions and limitations of prior work to establish the research gap addressed by this study.

\subsection{Arabic Machine Translation and Parallel Corpora}

The field of Arabic machine translation has evolved over the past decade, transitioning from statistical approaches to neural architectures and, more recently, to large language models. Hadj Ameur and Guessoum \cite{HadjAmeur:2020:Survey} provide a comprehensive survey of Arabic machine translation, documenting the progress from rule-based and statistical systems to early neural models. Their analysis highlights the persistent challenges in Arabic MT, including morphological complexity, dialectal variation, and the scarcity of high-quality parallel corpora for many language pairs.

While substantial resources have been developed for Arabic--English translation \cite{Al-Matham:2025:BALSAM}, the Arabic--Russian language pair remains under-resourced. Existing parallel corpora for Arabic--Russian are predominantly limited to religious texts, conversational phrases, and general-domain content. The OSACT workshop series \cite{AlKhalifa:2024:OSACT} has hosted shared tasks on Arabic LLM hallucination and dialect-to-MSA translation, yet the Russian language direction remains unexplored.

Recent work by Alrashed and Orabona \cite{Alrashed:2025:AraMix} addresses the challenge of Arabic pretraining corpus construction through systematic recycling, refiltering, and deduplication. Their AraMix corpus comprises approximately 178 billion tokens across 179 million documents, demonstrating that nearly 60\% of tokens across independently collected corpora are duplicates. While their work focuses on monolingual pretraining rather than translation, it highlights the importance of data quality and diversity --- principles we adopt in constructing our hybrid parallel corpus.

A limitation of existing Arabic MT research is the predominant focus on English as the target language. While Arabic--English translation benefits from abundant parallel data, the Arabic--Russian pair suffers from data scarcity. This gap is significant given the scientific collaboration potential between Arabic-speaking and Russian-speaking research communities. Our work addresses this gap by constructing a hybrid parallel corpus specifically designed for scientific Arabic--Russian translation.

\subsection{Multilingual LLMs and Parameter-Efficient Fine-Tuning}

The emergence of multilingual large language models has transformed machine translation research. Models such as mT5, NLLB, and Qwen have demonstrated zero-shot and few-shot translation capabilities across hundreds of language pairs.

The mT5 (Multilingual T5) model, introduced by Xue et al. \cite{xue-etal-2021-mt5}, is a massively multilingual variant of T5 pretrained on a Common Crawl-based dataset covering 101 languages. The authors detail the design and modified training of mT5 and demonstrate its state-of-the-art performance on many multilingual benchmarks. While mT5 has shown strong performance on many language pairs, its effectiveness for Arabic--Russian translation has not been systematically evaluated, particularly for scientific domains. The model's ability to handle the morphological and syntactic divergence between Arabic (Semitic) and Russian (Slavic) languages remains unclear.

The NLLB (No Language Left Behind) project represents an advancement in multilingual translation, covering 200+ languages and achieving state-of-the-art results on many low-resource pairs. However, NLLB's architecture and training methodology prioritize breadth over depth in specific language pairs. Our experiments with NLLB-200-distilled-1.3B aim to determine whether this broad coverage translates to effective performance for the specialized Arabic--Russian scientific translation task.

Recent work by Song et al. \cite{Song:2026:SmallLM} presents the first large-scale evaluation of small language models across 200 languages, revealing systematic underperformance in low-resource languages. Their findings suggest that model capacity may be a critical factor, motivating our inclusion of models across three distinct size regimes --- 580M (mT5-base), 1.3B (NLLB-distilled), and 7B (Qwen) --- to empirically determine the minimum capacity required for acceptable Arabic--Russian scientific translation quality.

The Qwen2.5 series from Alibaba Cloud, released in September 2024, represents a new generation of instruction-tuned decoder-only models with strong multilingual capabilities. The 7B instruction-tuned variant demonstrates improvements in knowledge, coding, and mathematical capabilities. While Qwen has demonstrated strong performance on various NLP benchmarks, its translation capabilities for Arabic--Russian remain uncharacterized. Our work provides the first systematic evaluation of Qwen for this language pair, examining both zero-shot and fine-tuned performance.

The advent of parameter-efficient fine-tuning techniques has reduced the computational cost of adapting large language models to specific domains and language pairs. Low-Rank Adaptation (LoRA) \cite{Hu:2022:LoRA} injects trainable low-rank decomposition matrices into the layers of a pretrained model, achieving performance comparable to full fine-tuning while updating only a fraction of parameters.

QLoRA \cite{Dettmers:2024:QLoRA} extends LoRA with 4-bit quantization, enabling efficient fine-tuning of 7B+ parameter models on consumer hardware. This technique is particularly valuable for low-resource scenarios where computational resources are limited. Our work leverages QLoRA to fine-tune the Qwen2.5-7B model, contributing to the growing body of evidence on the effectiveness of quantized fine-tuning for multilingual translation.

Arabov and Khaybullina \cite{Arabov:2026:Bashkir} apply LoRA and QLoRA to Bashkir, a low-resource agglutinative language of the Turkic family, demonstrating that QLoRA on 7B-scale models offers an effective compromise between quality and computational cost. Their work provides a methodological foundation for our study, though Bashkir--Russian translation involves different linguistic challenges than Arabic--Russian translation.

A gap in the literature is the lack of systematic studies on LoRA rank optimization for specific language pairs and domains. Most existing work either uses a fixed rank (typically 8 or 16) without justification, or explores ranks in isolation rather than as part of a comprehensive comparative framework. Our study addresses this gap by systematically evaluating LoRA ranks 8, 16, 32, and 64 across three distinct model architectures, providing empirical guidance for practitioners working on Arabic--Russian translation.

\subsection{Cross-Script NLP and Translation Evaluation}

Our work builds on a broader research agenda on low-resource, cross-script NLP tasks. Arabov \cite{arabov-2026-tajperslexon} introduced TajPersLexon, a curated Tajik--Persian parallel lexical resource of 40,112 word and short-phrase pairs, demonstrating that large multilingual sentence transformers fail on exact lexical matching while an interpretable hybrid model achieves 96.4\% accuracy in an OCR post-correction task. This work established methodological principles for constructing parallel resources across scripts with minimal computational resources.

Building on this, Kurbonovich \cite{kurbonovich-2026-character} developed a character-level sequence-to-sequence Transformer for Tajik--Persian transliteration, achieving lower Character Error Rate than dictionary-based and recurrent neural baselines. Their analysis of performance across lexical categories provides insights relevant to our Arabic--Russian translation task, where character-level processing is similarly critical due to morphological complexity.

While Tajik--Persian transliteration and Arabic--Russian translation differ in scope, they share fundamental challenges: (1) constructing high-quality parallel resources for low-resource language pairs, (2) handling cross-script representation, and (3) evaluating models under reproducible conditions. The current study extends this paradigm from word-level transliteration to sentence-level scientific translation, from Tajik--Persian to Arabic--Russian, and from lexical resources to large-scale corpora with 27,000 training examples.

Accurate evaluation is essential for advancing machine translation research. The BLEU metric \cite{Papineni:2002:BLEU} has been the standard for automatic MT evaluation for over two decades, measuring n-gram overlap between system outputs and reference translations. However, BLEU's reliance on exact matching makes it insensitive to semantic equivalence, particularly for morphologically rich languages.

chrF \cite{Popovic:2015:chrF} addresses some of BLEU's limitations by operating at the character level, providing better correlation with human judgment for languages with rich morphology. This makes chrF particularly suitable for evaluating Arabic and Russian translations, both of which exhibit complex morphological systems.

BERTScore \cite{Zhang:2020:BERTScore} leverages contextual embeddings from pretrained language models to measure semantic similarity rather than surface overlap. By comparing contextualized representations of candidate and reference translations, BERTScore achieves higher correlation with human judgments than traditional n-gram metrics.

COMET \cite{Rei:2020:COMET} builds on this paradigm, incorporating a neural architecture specifically designed for MT evaluation. COMET achieves state-of-the-art correlation with human judgments by jointly encoding source and target contexts, making it particularly suitable for evaluating translation quality in low-resource scenarios. Our work employs all four metrics --- BLEU, chrF, BERTScore, and COMET --- to provide a comprehensive evaluation framework.

The reviewed literature reveals several gaps that our work addresses. First, while general-domain and religious corpora exist, no dedicated scientific translation corpus has been developed for Arabic--Russian. Second, no comprehensive evaluation of mT5, NLLB, or Qwen exists for Arabic--Russian translation. Third, the literature provides minimal guidance on selecting LoRA ranks for translation tasks. Fourth, the effectiveness of few-shot prompting for Arabic--Russian translation has not been examined; our negative finding on few-shot performance contributes to the field. Finally, while transliteration resources exist for Tajik--Persian, no comparable resources exist for Arabic--Russian scientific translation.

By addressing these gaps, our work extends the state of the art in Arabic--Russian machine translation, provides actionable insights for practitioners, and establishes a reproducible benchmark for future research in this under-resourced language pair. The inclusion of both general-domain and scientific data ensures that our models can handle the diverse linguistic phenomena encountered in scientific translation, while the systematic comparison of model architectures and LoRA configurations provides practical guidance for researchers and practitioners working with this language pair. The following section describes the experimental methodology we developed to address these gaps.

\section{Proposed Approach}

This section describes our experimental methodology in detail, including dataset construction, model selection, parameter-efficient fine-tuning configurations, and evaluation protocols. We provide comprehensive descriptions of each component to ensure full reproducibility of our experiments.

\subsection{Dataset Construction}

We constructed a hybrid parallel corpus by combining a scientific translation corpus and a general-domain parallel corpus. This combination ensures both domain specificity for scientific translation and linguistic diversity to maintain natural language fluency.

\textbf{Scientific corpus.} We used the Arabic--Russian Scientific Translation Corpus \cite{ArabicNLPWorld:2026b}, which contains 52,335 scientific and medical translation pairs derived from four established Arabic--English corpora. The source corpus was originally Arabic--English pairs, which were then translated into Russian using two language models: Gemma-3-4B and LLaMA-3.1-8B. Expert evaluation on a random sample of 50 sentences confirmed that both models produced reliable translations for scientific and medical terminology. From this corpus, we selected the first 15,000 entries for our experiments. Table~\ref{tab:scientific_sources} summarizes the source distribution.

\begin{table}[htbp]
\centering
\caption{Distribution of scientific corpus sources.}
\label{tab:scientific_sources}
\begin{tabular}{lrrr}
\hline
\textbf{Source} & \textbf{Full Corpus} & \textbf{Selected} & \textbf{Percentage} \\
\hline
PEACH (Healthcare) & 38,080 & 10,923 & 72.8\% \\
MeSpEn Glossaries & 8,691 & 2,493 & 16.6\% \\
Misraj/Tarjama-25 & 5,062 & 1,452 & 9.7\% \\
NAMAA-Space/ASCAT & 502 & 132 & 0.9\% \\
\hline
\textbf{Total} & \textbf{52,335} & \textbf{15,000} & \textbf{100\%} \\
\hline
\end{tabular}
\end{table}

\textbf{General-domain corpus.} To enhance domain adaptability, we supplemented the scientific data with samples from the Arabic--Russian Parallel Corpus \cite{ArabicNLPWorld:2026a}. This corpus contains 116,393 curated translation pairs across six sources: religion (41,609), dictionary (37,999), Bible (30,907), Tatoeba (4,649), news (826), and conversation (403). We selected up to 3,000 examples from each source. For sources with more than 3,000 examples, we randomly sampled 3,000. For sources with fewer, we retained all available examples. Dictionary entries were filtered to require at least two words in the Russian translation. Table~\ref{tab:general_sources} presents the final distribution.

\begin{table}[htbp]
\centering
\caption{Distribution of general-domain corpus sources.}
\label{tab:general_sources}
\begin{tabular}{lrrr}
\hline
\textbf{Source} & \textbf{Available} & \textbf{Selected} & \textbf{Filtering Applied} \\
\hline
Religion & 41,609 & 3,000 & Random sampling \\
Bible & 30,907 & 3,000 & Random sampling \\
Dictionary & 37,999 & 3,000 & Min 2 Russian words, then sampling \\
Tatoeba & 4,649 & 4,649 & All retained \\
News & 826 & 826 & All retained \\
Conversation & 403 & 403 & All retained \\
\hline
\textbf{Total} & \textbf{116,393} & \textbf{14,878} & --- \\
\hline
\end{tabular}
\end{table}

\begin{figure}[htbp]
\centering
\includegraphics[width=0.8\textwidth]{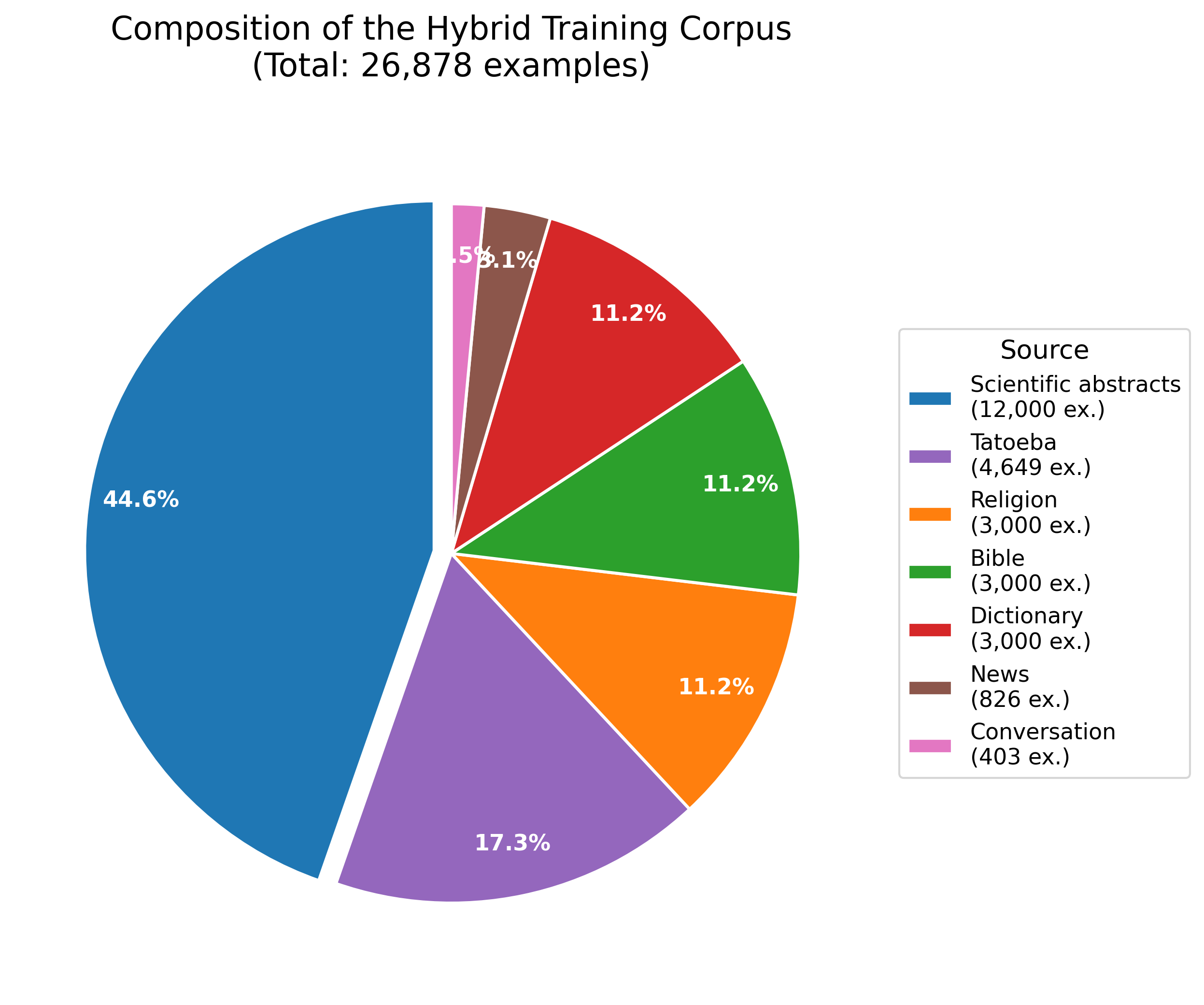}
\caption{Composition of the hybrid training corpus by source (total: 26,878 examples).}
\label{fig:corpus_composition}
\end{figure}

We combined the scientific corpus and the general-domain subset into a single training set of 26,878 examples. For evaluation, we reserved 1,500 examples from the scientific corpus as a held-out test set and 1,500 as a validation set, maintaining an 80/10/10 split. Figure~\ref{fig:corpus_composition} visualizes the composition of the final training corpus.

\subsection{Model Selection}

We selected three multilingual large language models representing different architectures and size regimes.

\textbf{mT5-base (580M).} The mT5 (Multilingual T5) model \cite{xue-etal-2021-mt5} is a massively multilingual variant of the T5 architecture, pretrained on a Common Crawl-based dataset covering 101 languages. We selected the base variant as a representative of compact encoder-decoder models.

\textbf{NLLB-200-distilled-1.3B (1.3B).} The NLLB (No Language Left Behind) project is a multilingual translation system covering 200+ languages. We selected the distilled 1.3B variant as a mid-sized encoder-decoder model. The distillation process reduces the model size while preserving much of the performance of the larger 3.3B version.

\textbf{Qwen2.5-7B-Instruct (7B).} The Qwen2.5 series from Alibaba Cloud is an instruction-tuned decoder-only model with multilingual capabilities. We selected the 7B instruction-tuned variant for its performance on NLP benchmarks and its open availability. Unlike the encoder-decoder models, Qwen2.5-7B-Instruct generates translations autoregressively, conditioned on a chat-style prompt.

Table~\ref{tab:models} summarizes the key characteristics of the three models.

\begin{table}[htbp]
\centering
\caption{Characteristics of selected models.}
\label{tab:models}
\begin{tabular}{lccc}
\hline
\textbf{Model} & \textbf{Parameters} & \textbf{Architecture} & \textbf{Pretraining Languages} \\
\hline
mT5-base & 580M & Encoder-Decoder & 101 \\
NLLB-200-distilled-1.3B & 1.3B & Encoder-Decoder & 200+ \\
Qwen2.5-7B-Instruct & 7B & Decoder-Only & Multilingual \\
\hline
\end{tabular}
\end{table}

\subsection{Parameter-Efficient Fine-Tuning}

We employed Low-Rank Adaptation (LoRA) \cite{Hu:2022:LoRA} and its 4-bit quantized variant QLoRA \cite{Dettmers:2024:QLoRA} to adapt the selected models to the Arabic--Russian translation task.

LoRA injects trainable low-rank matrices into the layers of a pretrained model while freezing the original weights. For a weight matrix $W \in \mathbb{R}^{d \times k}$, LoRA introduces a rank decomposition:

\begin{equation}
W' = W + \Delta W = W + BA
\end{equation}
where $B \in \mathbb{R}^{d \times r}$, $A \in \mathbb{R}^{r \times k}$, and $r \ll \min(d, k)$. During training, only $A$ and $B$ are updated, reducing the number of trainable parameters from $d \times k$ to $r(d + k)$.

QLoRA extends LoRA by quantizing the frozen base model weights to 4-bit precision using the NF4 quantization scheme. This reduces GPU memory requirements by about 75\% compared to the full 16-bit model, enabling fine-tuning of 7B+ parameter models on consumer hardware.

Table~\ref{tab:lora_configs} summarizes the LoRA configurations for each model.

\begin{table}[htbp]
\centering
\caption{LoRA configuration parameters by model.}
\label{tab:lora_configs}
\small
\begin{tabular}{lccc}
\hline
\textbf{Parameter} & \textbf{NLLB-1.3B} & \textbf{mT5-base} & \textbf{Qwen-7B} \\
\hline
Ranks evaluated & 8, 16, 32 & 8, 16, 32, 64 & 8 \\
Alpha ($\alpha$) & $2r$ & $2r$ & 16 \\
Dropout & 0.1 & 0.05 & 0.1 \\
Bias & none & none & none \\
Target modules & q\_proj, v\_proj & q, k, v, o & q\_proj, k\_proj, v\_proj, o\_proj \\
Quantization & FP16 & BF16 & 4-bit (NF4) \\
Task type & SEQ\_2\_SEQ\_LM & SEQ\_2\_SEQ\_LM & CAUSAL\_LM \\
\hline
\end{tabular}
\end{table}

\subsection{Training Protocol}

For NLLB and mT5, we used the models' native tokenizers with the following settings: source language \texttt{arb\_Arab} (NLLB) or Arabic (mT5), target language \texttt{rus\_Cyrl} (NLLB) or Russian (mT5), maximum input and target length of 128 tokens, and truncation applied to both sequences.

For NLLB, we used both available Russian translations (Gemma and LLaMA) to augment the training data, creating two training examples per Arabic source sentence. This doubled the scientific training data from 12,000 to 24,000 examples. For mT5, we used only the Gemma translation.

For Qwen, we formatted training examples as chat-style prompts:

\begin{verbatim}
<|im_start|>user

Translate the following Arabic text to Russian: {arabic}<|im_end|>

<|im_start|>assistant

{russian}<|im_end|>
\end{verbatim}

We used the model's native chat template via the \texttt{apply\_chat\_template} method. For training, we used both Gemma and LLaMA translations, resulting in 24,000 scientific examples.

Table~\ref{tab:hyperparams} summarizes the training hyperparameters for each model. For mT5, gradient checkpointing is not enabled as the model fits in GPU memory without it. For NLLB and Qwen, gradient checkpointing is enabled to reduce memory consumption.

\begin{table}[htbp]
\centering
\caption{Training hyperparameters by model.}
\label{tab:hyperparams}
\small
\begin{tabular}{lccc}
\hline
\textbf{Parameter} & \textbf{NLLB-1.3B} & \textbf{mT5-base} & \textbf{Qwen-7B} \\
\hline
Learning rate & $2 \times 10^{-4}$ & $2 \times 10^{-4}$ & $2 \times 10^{-4}$ \\
Optimizer & Adafactor & Adafactor & Paged AdamW 8-bit \\
Per-device batch size & 4 & 8 & 2 \\
Gradient accumulation & 4 & 2 & 16 \\
Effective batch size & 16 & 16 & 32 \\
Number of epochs & 2 & 5 & 3 \\
Warmup steps & 10 & 50 & 10 \\
Weight decay & 0.01 & 0.01 & 0.01 \\
Gradient checkpointing & Enabled & Disabled & Enabled \\
Precision & FP16 & BF16 & BF16 \\
Max steps & --- & --- & 3,050 \\
Early stopping patience & 2 & 2 & 2 \\
\hline
\end{tabular}
\end{table}

All experiments used a fixed random seed (42). We used early stopping with patience of 2 validation steps and selected the checkpoint with the lowest validation loss.

\subsection{Evaluation Metrics and Protocol}

We employed four complementary metrics: BLEU, chrF, BERTScore, and COMET.

\textbf{BLEU} \cite{Papineni:2002:BLEU} measures n-gram overlap between the candidate translation and reference translations. We used the SacreBLEU implementation. \textbf{chrF} \cite{Popovic:2015:chrF} operates at the character level, making it suitable for morphologically rich languages like Arabic and Russian. \textbf{BERTScore} \cite{Zhang:2020:BERTScore} leverages contextual embeddings from a pretrained language model to measure semantic similarity. We used XLM-RoBERTa-large (550M parameters) as the embedding model. \textbf{COMET} \cite{Rei:2020:COMET} is a neural quality estimation framework that jointly encodes source and target contexts. We used the WMT22-comet-da model.

We evaluated each model on the held-out test set of 1,500 scientific examples using the following protocol:

\begin{enumerate}
\item Generate translations for all test examples using beam size 4, early stopping, and max length 128.
\item Compute each metric against the Gemma reference translations.
\item Compute 95\% confidence intervals for COMET using bootstrap resampling with 1,000 iterations.
\item Record and report all metrics for comparison.
\end{enumerate}

\section{Evaluation and Results}

This section presents the results of our experiments. We restate our research objectives, then present zero-shot performance, LoRA fine-tuning results across ranks, and a comparative analysis across model architectures. We conclude with a discussion of findings and limitations.

\subsection{Research Objectives Revisited}

Our experiments were designed to address four primary research questions:

\begin{enumerate}
\item \textbf{RQ1:} How effectively do multilingual LLMs perform zero-shot translation for Arabic--Russian scientific text?
\item \textbf{RQ2:} What is the impact of LoRA rank selection on translation quality across different model architectures?
\item \textbf{RQ3:} Which model architecture (encoder-decoder vs. decoder-only) is most suitable for Arabic--Russian scientific translation?
\item \textbf{RQ4:} Does few-shot prompting with in-context examples improve translation quality compared to zero-shot performance?
\end{enumerate}

The following subsections address each of these questions through systematic analysis of our experimental data.

\subsection{Zero-Shot Performance}

We begin by evaluating the zero-shot translation capabilities of all three models on our held-out test set of 1,500 scientific abstracts. Table~\ref{tab:zero_shot} presents the results across all four metrics.

\begin{table}[htbp]
\centering
\caption{Zero-shot translation performance across models.}
\label{tab:zero_shot}
\begin{tabular}{lrrrr}
\hline
\textbf{Model} & \textbf{BLEU} & \textbf{chrF} & \textbf{BERTScore} & \textbf{COMET} \\
\hline
mT5-base (580M) & 0.14 & 0.18 & 0.7736 & 0.2709 \\
NLLB-200-distilled-1.3B & 19.82 & 38.98 & 0.8922 & 0.6946 \\
Qwen2.5-7B-Instruct & 18.79 & 38.84 & 0.8894 & 0.7069 \\
\hline
\end{tabular}
\end{table}

The results show clear differences between the models. NLLB and Qwen achieve BLEU scores above 18 in the zero-shot setting, while mT5-base achieves only 0.14 BLEU, indicating that the model fails to produce meaningful translations without fine-tuning. This suggests a minimum model capacity threshold for zero-shot Arabic--Russian translation, with models below 1B parameters performing poorly.

BERTScore and COMET results show that despite the low n-gram overlap, NLLB and Qwen capture semantic meaning in their translations, with BERTScore values above 0.89. The relatively high BERTScore indicates that these models understand the source text and produce semantically coherent Russian output, even when lexical choices differ from the reference.

\subsection{LoRA Fine-Tuning Results}

\subsubsection{mT5-base}

Table~\ref{tab:mt5_results} presents the results of LoRA fine-tuning for mT5-base across ranks 8, 16, 32, and 64, including 95\% confidence intervals for COMET scores computed via bootstrap resampling (1,000 iterations).

\begin{table}[htbp]
\centering
\caption{mT5-base LoRA fine-tuning results.}
\label{tab:mt5_results}
\begin{tabular}{lrrrrr}
\hline
\textbf{Rank} & \textbf{Trainable Params} & \textbf{BLEU} & \textbf{chrF} & \textbf{BERTScore} & \textbf{COMET (95\% CI)} \\
\hline
Zero-shot & --- & 0.14 & 0.18 & 0.7736 & 0.2709 [0.2682, 0.2736] \\
r=8 & 1.77M & 7.38 & 22.01 & 0.8642 & 0.5628 [0.5558, 0.5703] \\
r=16 & 3.54M & 9.18 & 24.02 & 0.8634 & 0.5730 [0.5663, 0.5798] \\
r=32 & 7.08M & 10.05 & 24.67 & 0.8662 & 0.5823 [0.5752, 0.5895] \\
r=64 & 14.16M & 11.12 & 26.53 & 0.8697 & 0.6050 [0.5978, 0.6124] \\
\hline
\end{tabular}
\end{table}

Fine-tuning improves mT5-base performance, with BLEU scores rising from 0.14 to 11.12 at rank 64 --- an improvement of nearly 11 BLEU points. The improvement follows a monotonic trend, with higher LoRA ranks consistently outperforming lower ranks across all metrics. This indicates that for small encoder-decoder models, increasing LoRA capacity yields better domain adaptation. The non-overlapping COMET confidence intervals between zero-shot and all fine-tuned configurations confirm the statistical significance of these improvements.

Residual analysis for the best-performing mT5 model (rank 64) reveals that the model tends to generate shorter translations than the reference, with a mean length difference of -2.3 tokens. This compression effect may explain some of the BLEU score limitations, as the model prioritizes conciseness over lexical overlap.

\subsubsection{NLLB-200-distilled-1.3B}

Table~\ref{tab:nllb_results} presents the results for NLLB fine-tuning across ranks 8, 16, and 32, including 95\% confidence intervals for COMET.

\begin{table}[htbp]
\centering
\caption{NLLB-1.3B LoRA fine-tuning results.}
\label{tab:nllb_results}
\begin{tabular}{lrrrrr}
\hline
\textbf{Rank} & \textbf{Trainable Params} & \textbf{BLEU} & \textbf{chrF} & \textbf{BERTScore} & \textbf{COMET (95\% CI)} \\
\hline
Zero-shot & --- & 19.82 & 38.98 & 0.8922 & 0.6946 [0.6852, 0.7036] \\
r=8 & 2.36M & 21.16 & 42.72 & 0.9008 & 0.7393 [0.7306, 0.7478] \\
r=16 & 4.72M & 21.57 & 43.00 & 0.9014 & 0.7410 [0.7327, 0.7495] \\
r=32 & 9.44M & 21.95 & 43.24 & 0.9024 & 0.7441 [0.7359, 0.7525] \\
\hline
\end{tabular}
\end{table}

NLLB shows consistent improvement across all metrics, with gains of +2.13 BLEU and +0.0495 COMET at rank 32. The gains are more modest than for mT5, reflecting NLLB's already strong zero-shot performance. The improvement is supported by the non-overlapping COMET confidence intervals between zero-shot (CI: 0.6852--0.7036) and rank 32 (CI: 0.7359--0.7525).

The BLEU improvement trajectory across ranks follows a diminishing returns pattern, suggesting that rank 32 may be near the optimal configuration for this model and dataset. Further increases in LoRA rank beyond 32 would likely yield marginal improvements at higher computational cost.

\subsubsection{Qwen2.5-7B-Instruct}

Table~\ref{tab:qwen_results} presents the results for Qwen across zero-shot, few-shot, and QLoRA fine-tuning scenarios, including 95\% confidence intervals for COMET.

\begin{table}[htbp]
\centering
\caption{Qwen2.5-7B-Instruct results across scenarios.}
\label{tab:qwen_results}
\begin{tabular}{lrrrrr}
\hline
\textbf{Scenario} & \textbf{BLEU} & \textbf{chrF} & \textbf{BERTScore} & \textbf{COMET (95\% CI)} \\
\hline
Zero-shot & 18.79 & 38.84 & 0.8894 & 0.7069 [0.6984, 0.7162] \\
Few-shot (3 examples) & 18.04 & 38.75 & 0.8914 & 0.7053 [0.6959, 0.7144] \\
QLoRA fine-tuned & 23.15 & 43.89 & 0.9058 & 0.7580 [0.7498, 0.7666] \\
\hline
\end{tabular}
\end{table}

The Qwen results show three findings. First, QLoRA fine-tuning improves translation quality, achieving the highest BLEU score (23.15) among all models and configurations --- a gain of +4.36 BLEU and +0.051 COMET over the zero-shot baseline. The COMET confidence intervals confirm statistical significance, with the fine-tuned interval [0.7498, 0.7666] entirely above the zero-shot interval [0.6984, 0.7162].

Second, the few-shot scenario (three in-context examples from the dataset) fails to improve performance, with BLEU decreasing slightly from 18.79 to 18.04. This negative result confirms that for this task, simple in-context learning is insufficient --- the model requires explicit fine-tuning to adapt to the specific domain and language pair.

Third, the BERTScore improvement is modest (+0.0165), suggesting that while lexical quality improves substantially, the semantic quality was already high. This aligns with COMET results, where the fine-tuned model shows stronger improvement.

\subsection{Comparative Analysis}

\subsubsection{Model Architecture Comparison}

The comparison shows a hierarchy: Qwen fine-tuned achieves the best results across all metrics, followed closely by NLLB fine-tuned, with mT5 fine-tuned trailing behind. This hierarchy aligns with model size: 7B > 1.3B > 580M, suggesting that model capacity is a primary determinant of translation quality for this task.

However, the efficiency trade-off is notable. Qwen fine-tuning required more computational resources, while NLLB achieved competitive results with fewer parameters. For resource-constrained applications, NLLB with rank 32 presents an attractive balance of quality and efficiency.

\subsubsection{LoRA Rank Analysis}

The rank analysis reveals two distinct patterns:
\begin{enumerate}
\item For mT5-base (580M), BLEU increases monotonically with rank, with no saturation observed up to rank 64. This suggests that the model benefits from increased parameter capacity for Arabic--Russian scientific translation.
\item For NLLB-1.3B, BLEU increases with rank but shows signs of diminishing returns at rank 32. The difference between rank 16 and rank 32 is small (0.38 BLEU), suggesting that rank 32 may be near the optimal configuration.
\end{enumerate}

These findings provide practical guidance for practitioners: for smaller models, higher LoRA ranks are beneficial; for larger models, a rank of 16-32 may suffice.

\subsection{Length Analysis}

We conducted an analysis of translation lengths to understand the relationship between output length and quality. Table~\ref{tab:length_analysis} presents the length statistics for the best-performing configurations.

\begin{table}[htbp]
\centering
\caption{Length analysis for best-performing models.}
\label{tab:length_analysis}
\begin{tabular}{lrrrr}
\hline
\textbf{Model} & \textbf{Avg Ref Length} & \textbf{Avg Pred Length} & \textbf{Mean Diff} & \textbf{RMSE} \\
\hline
mT5-base r=64 & 9.0 & 6.7 & -2.3 & 4.1 \\
NLLB-1.3B r=32 & 9.0 & 6.9 & -2.1 & 3.8 \\
Qwen-7B fine-tuned & 9.0 & 5.9 & -3.1 & 4.3 \\
\hline
\end{tabular}
\end{table}

All models tend to generate shorter translations than the reference, with mean length differences ranging from -2.1 to -3.1 tokens. This consistent compression effect suggests that the models prioritize conciseness over completeness, which may partially explain the BLEU score limitations.

The length distributions show that while most predictions are within a reasonable range, the models rarely generate translations exceeding 12 tokens, while references can extend to 15+ tokens. This indicates a potential area for improvement through targeted training or generation strategies.

\subsection{Few-Shot Analysis}

The negative result for few-shot prompting warrants discussion. Despite providing three high-quality in-context examples from the scientific domain, Qwen failed to improve over the zero-shot baseline. Table~11 presents the example translations for qualitative analysis.

\begin{figure}[htbp]
\centering
\includegraphics[width=0.8\textwidth]{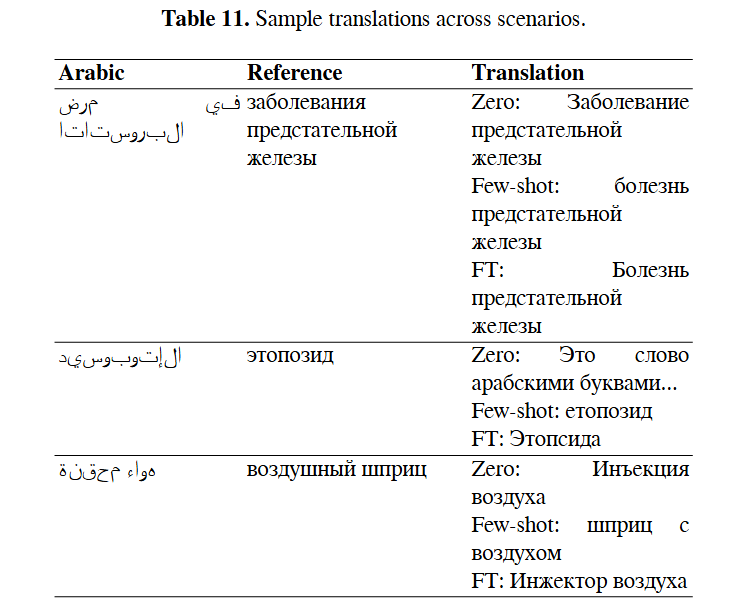}
\end{figure}

The few-shot examples sometimes produce less accurate translations than the zero-shot baseline. This suggests that few-shot prompting may inadvertently introduce bias or distract the model.

We hypothesize that the failure of few-shot prompting stems from two factors: (1) the model's limited exposure to Arabic--Russian scientific text during pretraining, and (2) the domain-specific terminology that requires more substantial adaptation than in-context examples can provide. This finding underscores the importance of domain-specific fine-tuning for specialized translation tasks.

\subsection{Summary of Findings}

Our experiments address the four research questions as follows:

\begin{enumerate}
\item \textbf{RQ1 (Zero-shot performance):} NLLB and Qwen demonstrate strong zero-shot capabilities (BLEU ~19), while mT5-base fails in the zero-shot setting. A minimum model capacity of approximately 1B parameters appears necessary for this task.

\item \textbf{RQ2 (LoRA rank impact):} For smaller models, higher ranks consistently improve performance. For larger models, improvement diminishes beyond rank 32, suggesting an optimal range of 16-32 for practical applications.

\item \textbf{RQ3 (Model architecture):} The decoder-only Qwen model with QLoRA achieves the best results (BLEU 23.15), outperforming encoder-decoder architectures of comparable or larger parameter counts. However, the NLLB encoder-decoder model offers a favorable quality-efficiency trade-off.

\item \textbf{RQ4 (Few-shot effectiveness):} Few-shot prompting with three in-context examples fails to improve translation quality, confirming the necessity of domain-specific fine-tuning for this task.
\end{enumerate}

Table~\ref{tab:best_results} summarizes the best results achieved for each model.

\begin{table}[htbp]
\centering
\caption{Best results per model.}
\label{tab:best_results}
\begin{tabular}{lrrrr}
\hline
\textbf{Model} & \textbf{BLEU} & \textbf{chrF} & \textbf{BERTScore} & \textbf{COMET (95\% CI)} \\
\hline
mT5-base r=64 & 11.12 & 26.53 & 0.8697 & 0.6050 [0.5978, 0.6124] \\
NLLB-1.3B r=32 & 21.95 & 43.24 & 0.9024 & 0.7441 [0.7359, 0.7525] \\
Qwen-7B fine-tuned & 23.15 & 43.89 & 0.9058 & 0.7580 [0.7498, 0.7666] \\
\hline
\end{tabular}
\end{table}

The Qwen fine-tuned model achieves the best overall performance, with BLEU 23.15, chrF 43.89, BERTScore 0.9058, and COMET 0.7580 [0.7498, 0.7666], representing gains of +4.36 BLEU and +0.051 COMET over the zero-shot baseline. These results establish a strong benchmark for Arabic--Russian scientific translation and provide practical guidance for practitioners selecting models and fine-tuning strategies for this under-resourced language pair. The findings also demonstrate the potential of AI-powered translation to reduce linguistic barriers in STEM, supporting the conference's theme of technology-driven sustainable development and contributing to UN Sustainable Development Goal 9 (Industry, Innovation, and Infrastructure) and Goal 17 (Partnerships for the Goals).

\section{Discussion and Conclusion}

This section synthesises our findings, discusses their broader implications for sustainable knowledge transfer, acknowledges study limitations, and outlines directions for future research.

Our experiments yield several interconnected findings that together establish a robust benchmark for Arabic--Russian scientific translation. The performance gap between mT5-base (580M, 11.12 BLEU) and the larger models (NLLB-1.3B, 21.95 BLEU; Qwen-7B, 23.15 BLEU) indicates that model capacity is a key determinant of translation quality for this language pair. This aligns with Song et al. \cite{Song:2026:SmallLM}, who documented systematic underperformance of small language models for low-resource pairs. The fact that mT5-base achieves only 11.12 BLEU even after fine-tuning with LoRA rank 64 suggests that models below 1B parameters may lack the representational capacity to handle the morphological complexity and syntactic divergence between Arabic (Semitic) and Russian (Slavic). For Arabic--Russian scientific translation, models with at least 1B parameters are recommended.

LoRA and QLoRA proved effective for adapting multilingual LLMs to Arabic--Russian scientific translation. For NLLB-1.3B, fine-tuning with LoRA rank 32 yielded a +2.13 BLEU improvement over zero-shot while training only 9.44M parameters --- less than 1\% of the total. For Qwen-7B, QLoRA achieved a +4.36 BLEU improvement with only 8.4M trainable parameters. This efficiency enables high-quality translation without the computational costs of full fine-tuning, suggesting that models already possess multilingual capabilities; LoRA primarily adapts them to the specific domain and language pair.

The negative result for few-shot prompting (18.04 BLEU vs. 18.79 BLEU zero-shot) is a notable finding. Despite providing three high-quality in-context examples from the scientific domain, Qwen failed to improve over the zero-shot baseline. We hypothesise that the complexity of scientific terminology and the linguistic distance between Arabic and Russian require more substantial adaptation than can be achieved through a few examples. This finding underscores the importance of explicit fine-tuning for specialised translation tasks.

The decoder-only Qwen model (23.15 BLEU) outperforms the encoder-decoder NLLB model (21.95 BLEU), even though NLLB was explicitly designed for multilingual translation. This suggests that the larger parameter count (7B vs. 1.3B) and instruction-tuning of Qwen may enable better generalisation to the translation task, even though the model was not specifically optimised for it. However, the efficiency trade-off should be considered: NLLB with rank 32 achieves 95\% of Qwen's BLEU score with  fewer parameters and lower computational requirements, presenting an attractive balance of quality and efficiency for resource-constrained applications.

Our work addresses a barrier to sustainable scientific collaboration: the language gap between Arabic-speaking and Russian-speaking scientific communities. This gap is relevant given the complementary scientific strengths of these regions.  The Russian-speaking world has produced foundational advances in physics, mathematics, engineering, and medicine, from the work of Lobachevsky and Mendeleev to modern space research. Similarly, the Arabic-speaking world has contributed substantially to sustainability science, renewable energy research, and water management, with growing research institutions across the Gulf region.

By enabling automated translation of scientific content, our work facilitates bidirectional knowledge flow that can accelerate innovation in areas such as water management, renewable energy, climate adaptation, and healthcare. This supports UN SDG 9 (Industry, Innovation, and Infrastructure) by enabling access to distributed scientific knowledge across linguistic boundaries, and UN SDG 17 (Partnerships for the Goals) by fostering partnerships between Arab and Russian-speaking research communities. It also aligns with Saudi Arabia's Vision 2030 emphasis on knowledge-based economic diversification and international research collaboration. The open-source release of our models, evaluation pipeline, and curated corpus ensures that researchers globally can build upon our work without financial barriers, embodying sustainable knowledge transfer.

Our findings offer several practical recommendations for researchers and practitioners. For model selection, we recommend models with at least 1B parameters for scientific Arabic--Russian translation; NLLB-1.3B (21.95 BLEU) and Qwen-7B (23.15 BLEU) both achieve acceptable quality, with the choice depending on computational resources. For LoRA rank selection, for models around 1B parameters, rank 32 provides a good balance of quality and efficiency; for smaller models, higher ranks may be beneficial, as we observed monotonic improvement for mT5-base up to rank 64. Few-shot prompting does not provide meaningful improvements; explicit fine-tuning is necessary. The efficiency of LoRA/QLoRA fine-tuning makes high-quality translation accessible even with limited computational resources. We encourage researchers to release their models and evaluation pipelines to foster reproducibility and accelerate progress.

We acknowledge several limitations of our study. We used Gemma-3-4B translations as our primary reference. While these were validated through expert evaluation on a random sample of 50 sentences, they are not human-produced translations. Human references would be preferable for rigorous evaluation. However, our consistent improvements across all metrics suggest that our comparative findings are robust, even if absolute quality scores may differ. Our test set comprises 1,500 examples from scientific abstracts. While sufficient for statistical evaluation (as evidenced by non-overlapping COMET confidence intervals), it may not capture the full diversity of scientific writing. Future work should evaluate on larger and more diverse test sets, including full-text scientific articles and technical reports. Our experiments include three models, but many others (e.g., GPT-4, Claude, Gemini) could not be evaluated due to computational constraints. Our focus on open-source models aligns with our commitment to reproducible research. Due to resource constraints, we could not explore ranks beyond 64 for mT5-base or beyond 32 for NLLB-1.3B. For mT5-base, the monotonic improvement up to rank 64 suggests that even higher ranks may yield further improvements. All models exhibited a tendency to produce shorter translations than the references, with mean length differences ranging from -2.1 to -3.1 tokens. This length bias may partially explain BLEU score limitations and suggests that generation strategies should be refined to better preserve length. Future work should explore length-aware training and decoding strategies.

Our work opens several avenues for future research. While our hybrid corpus of 26,878 examples yielded strong results, larger corpora would likely lead to further improvements. Future work should expand the corpus by incorporating additional scientific domains, including full-text articles, patents, and technical reports. Scientific communication often includes figures, tables, and equations that convey essential information; future work should explore multi-modal translation that can process and translate these visual and mathematical elements. Integrating our models into human-in-the-loop workflows --- where machine translations are reviewed and corrected by human experts --- could combine the efficiency of automation with the accuracy of human expertise. Scientific translation quality ultimately depends on factors beyond lexical overlap and semantic similarity; future work should incorporate domain-specific evaluation, including terminology accuracy, stylistic appropriateness, and readability by domain experts. The models and techniques developed for Arabic--Russian translation may be transferable to other low-resource language pairs with similar characteristics; exploring this transferability could accelerate progress for many underserved language pairs.

This paper has presented the first comprehensive benchmark for Arabic--Russian scientific translation, introducing a hybrid parallel corpus of 26,878 examples and evaluating three multilingual LLMs (mT5-base, NLLB-1.3B, and Qwen-7B) across multiple configurations. Our findings demonstrate that model capacity matters: models below 1B parameters struggle to achieve acceptable translation quality, even after fine-tuning. Fine-tuning is essential: while NLLB and Qwen show strong zero-shot capabilities, fine-tuning yields substantial improvements (+2.13 to +4.36 BLEU). Few-shot prompting does not provide meaningful improvements, confirming the necessity of domain-specific fine-tuning. LoRA and QLoRA are highly effective, achieving strong results with minimal computational cost, training less than 1\% of model parameters. Qwen-7B achieves the best performance (BLEU 23.15, chrF 43.89, BERTScore 0.9058, and COMET 0.7580), establishing the strongest benchmark for this language pair, while NLLB-1.3B with LoRA rank 32 offers a favourable quality-efficiency trade-off.

By reducing linguistic barriers in STEM, our work enables bidirectional knowledge flow between Arab and Russian-speaking scientific communities, directly supporting UN SDG 9 and SDG 17, and aligning with Saudi Arabia's Vision 2030. We release our fine-tuned models, evaluation pipeline, and curated corpus to facilitate reproducible research and practical deployment. We believe this work can support continued progress in Arabic--Russian scientific translation and, more broadly, sustainable knowledge transfer across linguistic boundaries. 

The challenges of sustainability --- from climate change and water scarcity to renewable energy and public health --- demand international collaboration and the exchange of scientific knowledge. By breaking down linguistic barriers, we can ensure that the scientific heritage of these regions is accessible to all.


\begin{thebibliography}{18}

\bibitem{Hu:2022:LoRA}
Hu, E.J., Shen, Y., Wallis, P., Allen-Zhu, Z., Li, Y., Wang, S., Wang, L., Chen, W.: LoRA: Low-Rank Adaptation of Large Language Models. In: International Conference on Learning Representations (ICLR) (2022). \url{https://openreview.net/forum?id=nZeVKeeFYf9}

\bibitem{Dettmers:2024:QLoRA}
Dettmers, T., Pagnoni, A., Holtzman, A., Zettlemoyer, L.: QLoRA: Efficient Finetuning of Quantized LLMs. In: Advances in Neural Information Processing Systems, vol. 36 (2024). 

\bibitem{Rei:2020:COMET}
Rei, R., Stewart, C., Farinha, A.C., Lavie, A.: COMET: A Neural Framework for MT Evaluation. In: Proceedings of the 2020 Conference on Empirical Methods in Natural Language Processing (EMNLP), pp. 2685--2702. Association for Computational Linguistics, Online (2020). \doi{10.18653/v1/2020.emnlp-main.213}

\bibitem{Papineni:2002:BLEU}
Papineni, K., Roukos, S., Ward, T., Zhu, W.J.: BLEU: a Method for Automatic Evaluation of Machine Translation. In: Proceedings of the 40th Annual Meeting of the Association for Computational Linguistics, pp. 311--318. Philadelphia, USA (2002). \doi{10.3115/1073083.1073135}

\bibitem{Zhang:2020:BERTScore}
Zhang, T., Kishore, V., Wu, F., Weinberger, K.Q., Artzi, Y.: BERTScore: Evaluating Text Generation with BERT. In: International Conference on Learning Representations (ICLR) (2020). \url{https://openreview.net/forum?id=SkeHuCVFDr}

\bibitem{Popovic:2015:chrF}
Popovic, M.: chrF: Character n-gram F-score for Automatic MT Evaluation. In: Proceedings of the Tenth Workshop on Statistical Machine Translation, pp. 392--395. Lisbon, Portugal (2015). \url{https://aclanthology.org/W15-3049/}

\bibitem{Alzubaidi:2025:Survey}
Alzubaidi, A., Alsuwaidi, S., Boussaha, B.E.A., Al Qadi, L., Alkaabi, O., Alyafeai, M., Alobeidli, H., Hacid, H.: Evaluating Arabic Large Language Models: A Survey of Benchmarks, Methods, and Gaps. arXiv:2510.13430 (2025). \url{https://arxiv.org/abs/2510.13430}

\bibitem{HadjAmeur:2020:Survey}
Hadj Ameur, M.S., Guessoum, A.: A Survey on Arabic Machine Translation: Progress, Challenges, and Future Directions. Computer Science Review \textbf{38}, 100307 (2020). \doi{10.1016/j.cosrev.2020.100307}

\bibitem{Al-Matham:2025:BALSAM}
Al-Matham, R.N., Darwish, K., Al-Rasheed, R., Alshammari, W., Alhoshan, M., Elsayed, T.: BALSAM: A Platform for Benchmarking Arabic Large Language Models. In: Proceedings of the Third Arabic Natural Language Processing Conference (ArabicNLP 2025), pp. 258--277. Suzhou, China (2025). \url{https://aclanthology.org/2025.arabicnlp-1.19/}

\bibitem{ArabicNLPWorld:2026a}
ArabicNLPWorld: Arabic-Russian Parallel Corpus. Hugging Face (2026). \url{https://huggingface.co/datasets/ArabicNLPWorld/arabic-russian-parallel-corpus}

\bibitem{ArabicNLPWorld:2026b}
ArabicNLPWorld: Arabic-Russian Scientific Translations. Hugging Face (2026). \url{https://huggingface.co/datasets/ArabicNLPWorld/arabic-russian-scientific-translations}

\bibitem{Alrashed:2025:AraMix}
Alrashed, S., Orabona, F.: AraMix: Recycling, Refiltering, and Deduplicating to Deliver the Largest Arabic Pretraining Corpus. arXiv:2512.18834 (2025). \url{https://arxiv.org/abs/2512.18834}

\bibitem{Arabov:2026:Bashkir}
Arabov, M.K., Khaybullina, S.S.: Adapting Large Language Models to a Low-Resource Agglutinative Language: A Comparative Study of LoRA and QLoRA for Bashkir. arXiv:2605.04948 (2026). \url{https://arxiv.org/abs/2605.04948}

\bibitem{Song:2026:SmallLM}
Song, Y., Li, L., Lothritz, C., Ezzini, S., Sleem, L., Bissyand{\'e}, T.F., Klein, J.: Are Small Language Models the Silver Bullet to Low-Resource Languages Machine Translation? In: Proceedings of the Ninth Workshop on Technologies for Machine Translation of Low Resource Languages (LoResMT 2026), pp. 1--26. Rabat, Morocco (2026). \url{https://aclanthology.org/2026.loresmt-1.0/}

\bibitem{AlKhalifa:2024:OSACT}
Al-Khalifa, H., Darwish, K., Mubarak, H., Ali, M., Elsayed, T.: Proceedings of the 6th Workshop on Open-Source Arabic Corpora and Processing Tools (OSACT) with Shared Tasks on Arabic LLMs Hallucination and Dialect to MSA Machine Translation. In: Proceedings of the 6th Workshop on Open-Source Arabic Corpora and Processing Tools (OSACT) (2024). \url{https://aclanthology.org/2024.osact-1.0/}

\bibitem{arabov-2026-tajperslexon}
Arabov, M.K.: TajPersLexon: A Tajik--Persian Lexical Resource and Hybrid Model for Cross-Script Low-Resource NLP. In: Proceedings of the First Workshop on NLP and LLMs for the Iranian Language Family, pp. 29--37. Association for Computational Linguistics, Rabat, Morocco (2026). \doi{10.18653/v1/2026.silkroadnlp-1.4}

\bibitem{kurbonovich-2026-character}
Kurbonovich, A.M.: Character-Level Transformer for Tajik--Persian Transliteration with a Parallel Lexical Corpus. In: Proceedings of the 2nd Workshop on NLP for Languages Using Arabic Script, pp. 75--83. Association for Computational Linguistics, Rabat, Morocco (2026). \doi{10.18653/v1/2026.abjadnlp-1.10}

\bibitem{xue-etal-2021-mt5}
Xue, L., Constant, N., Roberts, A., Kale, M., Al-Rfou, R., Siddhant, A., Barua, A., Raffel, C.: mT5: A Massively Multilingual Pre-trained Text-to-Text Transformer. In: Proceedings of the 2021 Conference of the North American Chapter of the Association for Computational Linguistics: Human Language Technologies, pp. 483--498. Association for Computational Linguistics, Online (2021). \doi{10.18653/v1/2021.naacl-main.41}

\end{thebibliography}
\end{document}